\documentclass[runningheads]{llncs}

\usepackage[mobile]{eccv}

\usepackage{eccvabbrv}

\usepackage{graphicx}
\usepackage{booktabs}
\usepackage{colortbl}
\usepackage{xcolor}
\usepackage{multirow}
\usepackage{multicol}
\usepackage{makecell}

\usepackage[accsupp]{axessibility}  %

\usepackage{hyperref}

\usepackage{orcidlink}

\newcommand{\model}{\textbf{DVGT-2}\xspace}

\begin{document}

\title{DVGT-2: Vision-Geometry-Action Model for Autonomous Driving at Scale}

\titlerunning{DVGT-2}

\author{Sicheng Zuo$^{1,*}$ \and
Zixun Xie$^{1,2,4,*}$ \and
Wenzhao Zheng$^{1,*, \dagger}$ \and
Shaoqing Xu$^{2, \dagger}$ \and
Fang Li$^{2}$ \and
Hanbing Li$^{2}$ \and
Long Chen$^{2}$ \and
Zhi-Xin Yang$^{3}$ \and
Jiwen Lu$^{1}$
}

\authorrunning{S. Zuo, Z. Xie, W. Zheng et al.}

\institute{$^1$Tsinghua University \samelineand
$^2$Xiaomi EV \samelineand
$^3$University of Macau \samelineand
$^4$Peking University \\
Project Page: \url{https://wzzheng.net/DVGT-2}\\
Large Driving Models: \url{https://github.com/wzzheng/LDM} \\
}
\newcommand{\samelineand}{\quad}

\maketitle

\begin{figure}[!h]
\centering
\vspace{-8mm}
\includegraphics[width=0.9\textwidth]{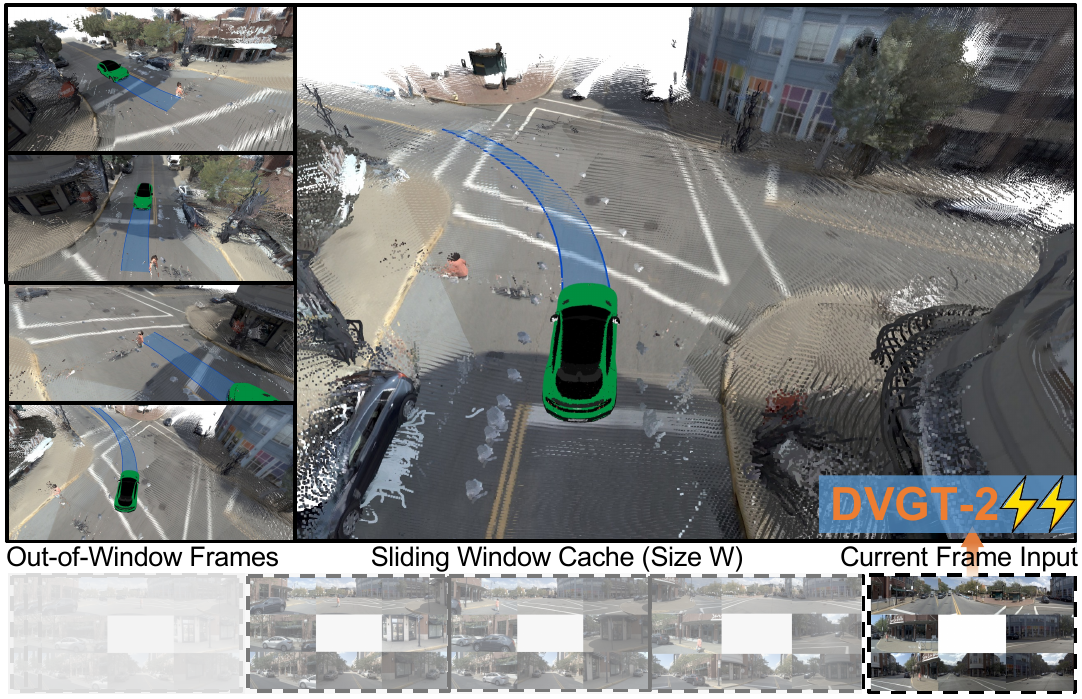}
\vspace{-3.5mm}
\caption{\textbf{\model is a streaming visual geometry transformer specifically designed for autonomous driving}. It inputs multi-view images and jointly predicts the 3D pointmaps, the ego poses, and future trajectory planning in an online manner.}
\label{fig:teaser}
\vspace{-12mm}
\end{figure}

\begingroup
\renewcommand\thefootnote{}
\footnotetext{
  $^*$Equal contributions. 
  $^\dagger$Project leaders.
}
\endgroup

\begin{abstract}
End-to-end autonomous driving has evolved from the conventional paradigm based on sparse perception into vision-language-action (VLA) models, which focus on learning language descriptions as an auxiliary task to facilitate planning.
In this paper, we propose an alternative Vision-Geometry-Action (VGA) paradigm that advocates dense 3D geometry as the critical cue for autonomous driving. 
As vehicles operate in a 3D world, we think dense 3D geometry provides the most comprehensive information for decision-making.
However, most existing geometry reconstruction methods (e.g., DVGT) rely on computationally expensive batch processing of multi-frame inputs and cannot be applied to online planning.
To address this, we introduce a streaming Driving Visual Geometry Transformer (\model), which processes inputs in an online manner and jointly outputs dense geometry and trajectory planning for the current frame.
We employ temporal causal attention and cache historical features to support on-the-fly inference.
To further enhance efficiency, we propose a sliding-window streaming strategy and use historical caches within a certain interval to avoid repetitive computations.
Despite the faster speed, \model achieves superior geometry reconstruction performance on various datasets.
The same trained \model can be directly applied to planning across diverse camera configurations without fine-tuning, including closed-loop NAVSIM and open-loop nuScenes benchmarks.
Code is available at \href{https://github.com/wzzheng/DVGT}{{\texttt{https://github.com/wzzheng/DVGT}}}.
\end{abstract}
    
\section{Introduction}
\label{sec: intro}
End-to-end autonomous driving has recently seen remarkable progress, fundamentally changing how vehicles perceive and navigate complex environments~\cite{uniad, vad, vadv2, transfuser, diffusiondrive}. 
Conventional methods typically rely on sparse perception representations (e.g., 3D object detection~\cite{bevformer, bevdet, sparse4d, sparse4dv2, sparse4dv3, xu2024sparseinteraction,xu2021fusionpainting,tigdistillbev} and map segmentation~\cite{bevformer, bevfusion, bevfusion-2,gaussianpretrain}) to guide scene understanding and trajectory planning. 
Recently, the emerging Vision-Language-Action (VLA) models leverage the general understanding and reasoning capabilities of pre-trained Vision-Language Models (VLMs), focusing on learning natural language descriptions to interpret driving scenarios and facilitate robust decision-making~\cite{omnidrive, emma, autovla, recogdrive,luo2025adathinkdrive}.

As vehicles inherently operate in a 3D world, we argue that understanding the dense 3D geometry of the environment provides the most direct and comprehensive information for safe decision-making.
Although language descriptions offer valuable high-level context, they can sometimes be too coarse to capture complete and accurate geometric details when precise spatial control is required.
Considering this, we explore an alternative Vision-Geometry-Action (VGA) paradigm for end-to-end autonomous driving.
Rather than rely on sparse perception representations or coarse language descriptions, VGA advocates for dense 3D geometry as the foundational representation for driving.
By explicitly recovering pixel-aligned 3D pointmaps, VGA extracts comprehensive and precise geometric cues from visual inputs, directly empowering trajectory planning.

However, integrating high-fidelity geometry reconstruction into an end-to-end planning framework presents significant computational challenges. 
Existing methods~\cite{vggt, pi3, dvgt} typically rely on computationally expensive batch processing of the entire multi-frame sequence.
When processing online inputs frame-by-frame, they have to recalculate features for overlapping historical frames at every time step.
Such redundant computation incurs unacceptable latency and makes them unsuitable for online, real-time autonomous driving.
To address this, we introduce a streaming Driving Visual Geometry Transformer (\model), designed to jointly output dense geometry and trajectory planning in an online manner, as shown in ~\cref{fig:teaser}. 
Instead of reprocessing the entire historical sequence, \model employs temporal causal attention and caches historical intermediate features to support efficient on-the-fly inference. 
To further enhance efficiency and bound the computational overhead for continuous driving, we propose a sliding-window streaming strategy that only utilizes historical caches within a fixed interval. 
We achieve this by introducing relative temporal positional encoding and jointly predicting dense points in the local coordinate system alongside the ego-pose relative to the previous frame. 
This design allows the model to continuously aggregate historical geometric cues and avoid repetitive computations. 

To build a robust foundation model for the VGA paradigm, we train \model on a large mixture of diverse driving datasets, including nuScenes~\cite{nuscenes}, OpenScene~\cite{openscene}, Waymo~\cite{waymo}, KITTI~\cite{kitti}, and DDAD~\cite{ddad}.
Extensive experiments demonstrate that, despite its significantly faster inference speed, our model achieves superior geometry reconstruction performance on various datasets.
More importantly, the same trained \model can be directly applied to trajectory planning across diverse datasets without the need for finetuning, achieving strong performance on both the closed-loop NAVSIM~\cite{navsim-v1,navsim-v2} and open-loop nuScenes~\cite{nuscenes} benchmarks, validating the effectiveness of the VGA paradigm.
  
\section{Related Work}
\label{sec: related work}

\textbf{End-to-end Autonomous Driving.}
End-to-end autonomous driving aims to directly map raw sensor inputs to planning trajectories or control signals, jointly optimizing the entire system to minimize error accumulation.
Pioneering works like UniAD~\cite{uniad} and VAD~\cite{vad} proposed the paradigm by integrating perception, prediction, and planning into a single framework, achieving planning-oriented joint optimization across all tasks. 
Subsequent research shifted towards multi-modal trajectory generation. 
VADv2~\cite{vadv2} and Hydra-MDP~\cite{hydra-mdp} modeled probabilistic planning by sampling from a fixed trajectory codebook. DiffusionDrive~\cite{diffusiondrive} proposed an anchor-based truncated diffusion model to capture the multi-modal distribution of trajectories, which is adopted by other methods~\cite{trajhf, transdiffuser, goalflow}. 
Despite these advances, existing methods~\cite{genad, gaussianad, vadv2} fundamentally depend on manually defined perception tasks, such as detection~\cite{bevformer,bevdet,bevdepth,sparse4d}, tracking~\cite{bevfusion, sparse4dv3, sparsead, streampetr}, or occupancy~\cite{tpvformer,gaussianformer,gaussianformer2,gaussianworld,quadricformer} to understand driving scenes, which is extremely inefficient with world information.
In contrast, our model explicitly reconstructs fine-grained dense geometry. By comprehensively modeling the geometric details, our approach facilitates robust trajectory planning.

\begin{figure}[!t]
\centering
\includegraphics[width=1.0\textwidth]{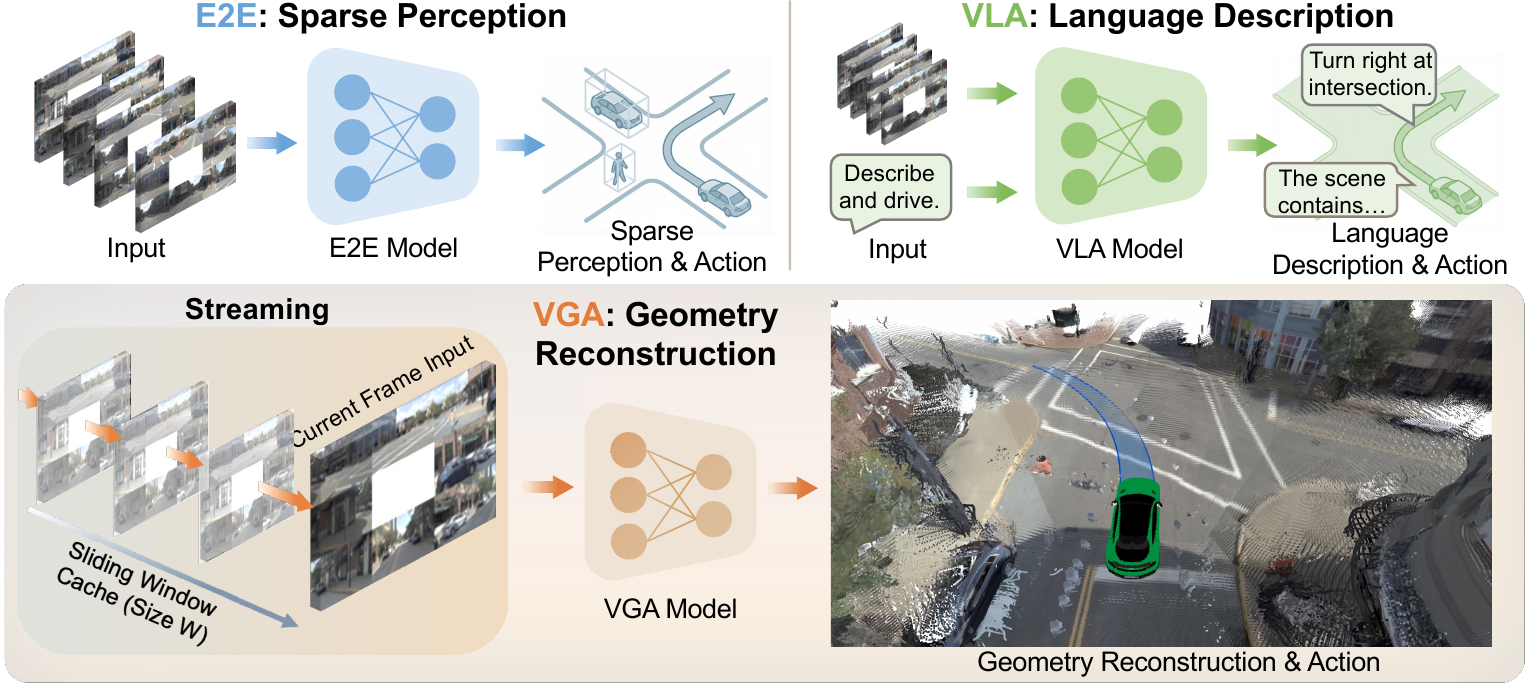}
\vspace{-7mm}
\caption{\textbf{Comparison of different paradigms for end-to-end autonomous driving}. Conventional end-to-end models rely on sparse perception representations for scene understanding. VLA models predict language descriptions to interpret driving scenarios. Our VGA model reconstructs dense 3D geometry to facilitate safe planning.}
\label{fig:vga}
\vspace{-6mm}
\end{figure}

\textbf{VLA for Autonomous Driving.}
The extensive world knowledge of Vision-Language Models (VLMs) have driven their applications in autonomous driving~\cite{lmdrive, drivemm, drivegpt4}, mainly focus on scene understanding and reasoning. 
Subsequent research utilized VLMs to predict high-level meta-actions or driving decisions~\cite{alphadrive, senna, drivemlm}, which serve as intermediate guidance~\cite{senna,cotdrive} or supervision~\cite{vlp, vlm-ad, dima, vlm-e2e} for downstream planners or end-to-end models. 
Although these methods facilitate the integration of VLM knowledge, the modular design hinders full end-to-end optimization.
Recent advancements have integrated planning modules into the VLMs to map sensor inputs to planning trajectories directly. 
Initial attempts sought to predict trajectories directly in text format~\cite{omnidrive, emma, luo2026unleashing, gpt-driver, wisead, luo2025adathinkdrive,luo2026last},
while other methods explored incorporating specialized trajectory decoders to predict feasible actions, including the auto-regressive manner~\cite{opendrivevla, diffvla, autovla}, the diffusion module~\cite{drivemoe, orion, recogdrive}, or the MLP head~\cite{simlingo, carllava, dsdrive}. 
In contrast, we propose a Vision-Geometry-Action (VGA) framework, as shown in~\cref{fig:vga}. By explicitly modeling fine-grained dense geometry, our approach connects visual inputs to driving actions through high-fidelity structural and dynamic scene understanding, thereby enhancing trajectory planning.

\textbf{Visual Geometry Models.}
Recent advances in general visual geometry models have shifted the paradigm from depth estimation to regressing dense pointmaps directly from unposed images. 
Pioneering works like DUSt3R~\cite{dust3r} and its extensions~\cite{mast3r, spann3r, point3r} have demonstrated impressive capabilities in recovering 3D structure from image pairs. 
More recently, large-scale geometry foundation models like VGGT~\cite{vggt} and $\pi^3$~\cite{pi3} have achieved robust geometry reconstruction across multi-view inputs. 
However, these methods are limited to a relative scale and require post-alignment with sparse LiDAR points to recover the metric-scaled scene geometry. 
To address this, DVGT~\cite{dvgt} proposed a driving visual geometry model to enable end-to-end reconstruction of driving scene geometry with metric scale. 
But it relies on batch processing of multi-frame inputs, which leads to redundant computations for historical frames.
While streaming alternatives like CUT3R~\cite{cut3r} and StreamVGGT~\cite{streamvggt} attempt to reduce latency, they typically maintain a feature cache that grows linearly with input length, making them unsuitable for infinite-length driving scenarios. 
In contrast, we propose a window-based streaming architecture by maintaining a fixed-size historical cache and predicting local geometry only for the latest frame. Our approach thus ensures constant computational costs and supports efficient, long-term inference.

\section{Proposed Approach}
\label{sec: method}

\subsection{Vision-Geometry-Action Model}
The fundamental objective of an autonomous driving system is to predict a safe and feasible future trajectory for the ego vehicle given historical and current sensor observations. 
Formally, at the current time step $t$, given a sequence of multi-view image inputs $\mathbf{I}_{t-T:t}$ from the past $T$ frames and the current frame, the driving model $\mathcal{M}$ aims to predict the future ego-trajectory $\mathbf{A}_t$:
\begin{equation}
    \mathbf{A}_t = \mathcal{M}(\mathbf{I}_{t-T:t}).
\end{equation}

Conventional end-to-end models typically decompose this process into a sequential perception-prediction-planning pipeline~\cite{uniad, vad, gaussianad}. 
The perception module $\mathcal{F}_{\text{perc}}$ first extracts high-level perceptual representations $\mathbf{Z}$ from the visual inputs. 
The prediction module $\mathcal{F}_{\text{pred}}$ then forecasts the future motion $\mathbf{V}$ of surrounding agents. 
Finally, the planning module $\mathcal{F}_{\text{plan}}$ utilizes these representations to generate the future ego-trajectory. 
This process can be expressed as:
\begin{align}
    \mathbf{Z} = \mathcal{F}_{\text{perc}}(\mathbf{I}_{t-T:t}), \ \ 
    \mathbf{V} = \mathcal{F}_{\text{pred}}(\mathbf{Z}), \ \ 
    \mathbf{A}_t = \mathcal{F}_{\text{plan}}(\mathbf{Z}, \mathbf{V}).
\end{align}
However, this paradigm heavily relies on sparse perceptual representations like bounding boxes and map elements, which discard rich environmental context. 
While some methods introduce 3D occupancy to provide a dense structural description, they inherently suffer from quantization errors due to voxel discretization. 
Consequently, this incomplete and inaccurate modeling of scene structures fundamentally restricts the model's planning performance.

Recently, Vision-Language-Action (VLA) models have emerged as a promising alternative, leveraging the strong semantic understanding capabilities of pre-trained Vision-Language-Models (VLMs) to facilitate driving~\cite{omnidrive, emma, doe-1, autovla}. 
Given historical observations, a VLA model $\mathcal{M}_{\text{VLA}}$ typically outputs a textual description $\mathbf{L}_t$ of the current scene alongside the future trajectory prediction $\mathbf{A}_t$:
\begin{equation}
    \mathbf{A}_t, \mathbf{L}_t = \mathcal{M}_{\text{VLA}}(\mathbf{I}_{t-T:t}).
\end{equation}
Although VLA models demonstrate remarkable generalization, natural language is inherently ambiguous and coarse-grained. 
It struggles to accurately and comprehensively capture the precise geometric details of driving scenes, thereby hindering high-fidelity scene understanding and robust trajectory planning.

In this paper, we propose a novel \textbf{Vision-Geometry-Action (VGA)} framework, which identifies dense geometry as the critical bridge connecting visual inputs to driving actions. 
Given multi-frame image inputs, our VGA model $\mathcal{M}_{\text{VGA}}$ jointly reconstructs the dense 3D pointmaps $\mathbf{P}_{t-T:t}$ and ego-poses $\mathbf{E}_{t-T:t}$, and predicts the future ego-trajectory $\mathbf{A}_t$:
\begin{equation}
    \mathbf{A}_t, \mathbf{P}_{t-T:t}, \mathbf{E}_{t-T:t} = \mathcal{M}_{\text{VGA}}(\mathbf{I}_{t-T:t}).
\end{equation}
This paradigm offers two fundamental advantages. 
First, the continuous coordinate space of dense pointmaps eliminates quantization errors, providing a pixel-aligned, complete representation of both foreground objects and background environments. 
Second, by explicitly modeling spatial geometry and camera poses across multi-frame inputs, the VGA model comprehensively captures temporally consistent static structures and coherent dynamic motions, thereby providing a more precise and reliable foundation for trajectory planning.

\subsection{Streaming Geometry Reconstruction}

\begin{figure}[!t]
\centering
\includegraphics[width=1.0\textwidth]{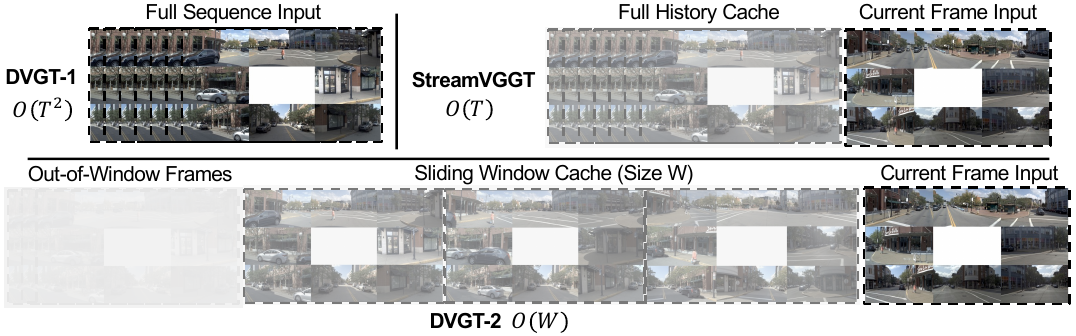}
\vspace{-7mm}
\caption{\textbf{Comparison of different paradigms for geometry reconstruction}. Batch-processing models like DVGT~\cite{dvgt} compute pair-wise relations across all frames, incurring an overall $\mathcal{O}(T^2)$ complexity. Full-history streaming models like StreamVGGT~\cite{streamvggt} extract temporal cues from the entire history, leading to an $\mathcal{O}(T)$ per-frame complexity. In contrast, our sliding-window streaming strategy attends to a fixed-size cache of length $W$, achieving a constant $\mathcal{O}(W)$ per-frame complexity.}
\label{fig:stream}
\vspace{-6mm}
\end{figure}

In the VGA framework, dense geometry reconstruction from multi-frame inputs is essential for comprehensive scene understanding.
Recently, numerous works~\cite{vggt, pi3, dvgt} have made progress by adopting a batch-processing paradigm. 
Given a sequence of image inputs $\mathbf{I}_{t-T:t}$, these models jointly reconstruct global pointmaps and ego-poses for all frames, typically anchored to the first frame's coordinate system:
\begin{equation}
    \mathbf{P}_{t-T:t}, \mathbf{E}_{t-T:t} = \mathcal{G}_{\text{batch}}(\mathbf{I}_{t-T:t}),
\end{equation}
where $\mathcal{G}_{\text{batch}}$ denotes batch-processing reconstruction paradigm.
However, this paradigm requires computing pairwise spatial interactions across all input frames, resulting in an $\mathcal{O}(T^2)$ computational complexity. 
More severely, when processing online inputs frame-by-frame in driving scenarios, the model redundantly reprocesses the overlapping historical frames at every time step.
This massive computational redundancy leads to prohibitive inference latency, making it entirely unsuited for online, real-time autonomous driving applications~\cite{dvgt}.

To alleviate this redundant computation, recent works like StreamVGGT~\cite{streamvggt} introduce a full-history streaming reconstruction framework with a feature caching mechanism. 
When receiving a new frame input $\mathbf{I}_{t}$, the model $\mathcal{G}_{\text{stream}}$ only computes the interaction between the current frame and the cached historical features $\mathbf{C}_{t-T:t-1}$, thereby reducing the computational complexity from $\mathcal{O}(T^2)$ to $\mathcal{O}(T)$. 
After the prediction, the current frame's features are updated into the cache for the next-frame prediction.
This process can be formulated as:
\begin{equation}
    \mathbf{P}_t, \mathbf{E}_t, \mathbf{C}_{t-T:t} = \mathcal{G}_{\text{stream}}([\mathbf{I}_{t}, \mathbf{C}_{t-T:t-1}]).
\end{equation}
Although this framework avoids repeatedly computing historical frames, it still relies on the first frame as the global reference coordinate system. 
Consequently, the model must retain features of the entire history. 
This causes memory and computational costs that scale linearly with the sequence length, rendering the approach prohibitive for continuous, infinite-horizon driving scenarios.     

To overcome these bottlenecks, we propose a sliding-window streaming strategy, as shown in~\cref{fig:stream}. 
The core idea is to maintain a historical feature cache $\mathcal{C}_{t-W:t-1}$ with a fixed window size of $W$, ensuring a constant $\mathcal{O}(W)$ per-frame complexity. 
Crucially, to eliminate the dependency on the full historical sequence, we decouple the geometry reconstruction from the first-frame coordinate system. 
Instead, we reconstruct the local geometry $\mathbf{P}_t$ in the current frame's ego-coordinate system and predict the ego-pose $\mathbf{E}_t$ relative to the previous frame. Our sliding-window streaming inference process can be formulated as:
\begin{equation}
    \mathbf{P}_t, \mathbf{E}_t, \mathbf{C}_{t-W+1:t} = \mathcal{G}_{\text{window}}([\mathbf{I}_{t}, \mathbf{C}_{t-W:t-1}]),
\end{equation}
where $\mathcal{G}_{\text{window}}$ denotes sliding-window streaming reconstruction paradigm.
After processing the current frame, the cache is updated in a First-In-First-Out (FIFO) manner, discarding the earliest frame's features $\mathbf{C}_{t-W}$ and appending the current frame's features $\mathbf{C}_{t}$. 
Our sliding-window streaming framework can efficiently process arbitrary-length video streams with constant overhead, adhering to the strict efficiency constraints of real-time autonomous driving systems.

\subsection{Streaming Driving Visual Geometry Transformer}

\begin{figure}[!t]
\centering
\includegraphics[width=1.0\textwidth]{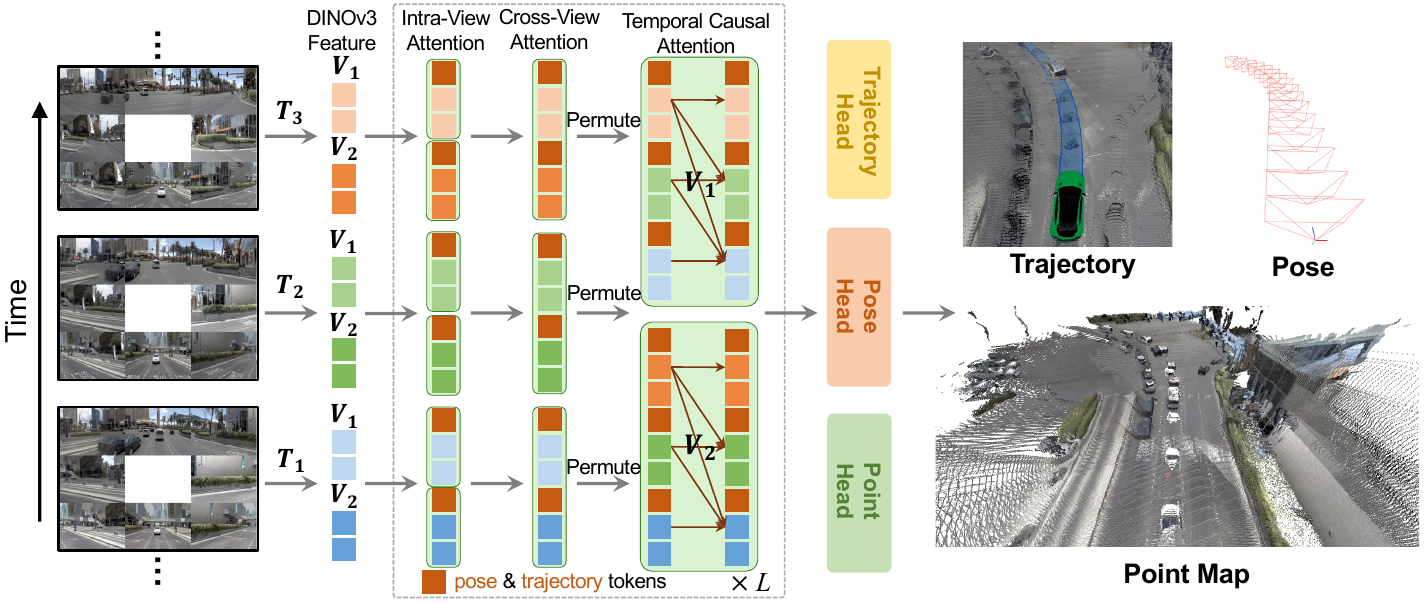}
\vspace{-7mm}
\caption{\textbf{Overall archetecture of \model}. Our model consists of an image encoder, a geometry transformer with temporal causal attention, and a set of prediction heads to jointly output geometry reconstruction and trajectory planning. }
\label{fig:arch}
\vspace{-6mm}
\end{figure}

To realize the sliding-window streaming paradigm for efficient geometry reconstruction and trajectory planning, we propose \model, a Streaming Driving Visual Geometry Transformer.
At time step $t$, given multi-view image inputs $\mathbf{I}_{t} \in \mathbb{R}^{V \times H \times W \times 3}$ from $V$ cameras and a historical feature cache $\mathbf{C}_{t-W:t-1}$ containing the past $W$ frames, the model predicts three components: 
(1) the multi-view 3D pointmaps $\mathbf{P}_t \in \mathbb{R}^{V\times H\times W\times 3}$ in the current ego-coordinate system; 
(2) the current ego-pose $\mathbf{E}_t \in \mathbb{R}^7$ (comprising a 3D translation and a 4D rotation quaternion) relative to the previous frame;
(3) the future $N$-step planning trajectory $\mathbf{A}_t \in \mathbb{R}^{N\times 3}$ (representing x, y coordinates, and yaw angle).
The cache is then updated to the current frame. 
The overall process can be expressed as:
\begin{equation}
    \mathbf{A}_t, \mathbf{P}_t, \mathbf{E}_t, \mathbf{C}_{t-W+1:t} = \mathcal{M}_{\text{DVGT-2}}(\mathbf{I}_{t}, \mathbf{C}_{t-W:t-1}).
\end{equation}

\begin{figure}[!t]
\centering
\includegraphics[width=1.0\textwidth]{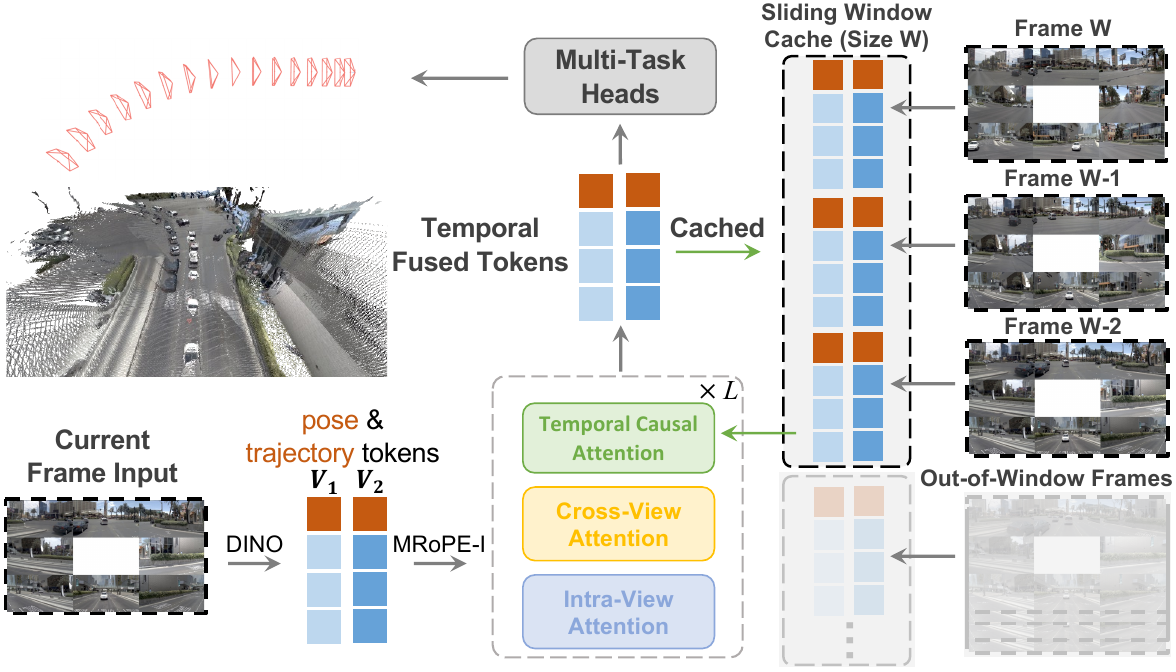}
\vspace{-7mm}
\caption{\textbf{Efficient inference of \model}. Given the current frame multi-view input and the cache of past $W$ frames, our model performs efficient geometry reconstruction and trajectory planning in an online manner, avoiding recomputing historical frames.}
\label{fig:infer}
\vspace{-6mm}
\end{figure}

\textbf{Overall Architecture}. 
As shown in ~\cref{fig:arch}, our \model consists of an image encoder $\mathcal{E}$, a geometry transformer $\mathcal{G}$, and task-specific prediction heads $[\mathcal{H}^{\text{vis}}, \mathcal{H}^{\text{pose}}, \mathcal{H}^{\text{traj}}]$. 
Given the multi-view images $\mathbf{I}_{t}$ of the current time step $t$, we first employ a pre-trained vision foundation model~\cite{dinov3} to extract visual tokens:
\begin{equation}
    \mathbf{F}_t^{\text{vis}} = \mathcal{E}(\mathbf{I}_t).
\end{equation}
To explicitly aggregate global representations for spatial reasoning and planning, we append learnable pose tokens $\mathbf{F}_t^{\text{pose}}$ and trajectory tokens $\mathbf{F}_t^{\text{traj}}$ to the visual tokens of each view, forming the unified input tokens:
\begin{equation}
    \mathbf{F}_t = [\mathbf{F}_t^{\text{vis}}, \mathbf{F}_t^{\text{pose}}, \mathbf{F}_t^{\text{traj}}].
\end{equation}
Subsequently, these combined tokens, along with the historical cache $\mathbf{C}_{t-W:t-1}$, are fed into the geometry transformer $\mathcal{G}$ for spatial-temporal reasoning:
\begin{equation}
    \mathbf{G}_t^{\text{vis}}, \mathbf{G}_t^{\text{pose}}, \mathbf{G}_t^{\text{traj}} = \mathcal{G}(\mathbf{F}_t, \mathbf{C}_{t-W:t-1}).
\end{equation}
Finally, the updated tokens are routed to their respective prediction heads to generate the 3D pointmaps $\mathbf{P}_t$, ego-pose $\mathbf{E}_t$, and future trajectory $\mathbf{A}_t$:
\begin{equation}
    \mathbf{P}_t = \mathcal{H}^{\text{vis}}(\mathbf{G}_t^{\text{vis}}), \ \
    \mathbf{E}_t = \mathcal{H}^{\text{pose}}(\mathbf{G}_t^{\text{pose}}), \ \
    \mathbf{A}_t = \mathcal{H}^{\text{traj}}(\mathbf{G}_t^{\text{traj}}).
\end{equation}
The overall pipeline of our model's efficient online inference is illustrated in~\cref{fig:infer}.

\textbf{Geometry Transformer.} 
Following DVGT~\cite{dvgt}, we utilize a factorized attention mechanism to efficiently model the complex spatial-temporal relations across multiple views and frames. 
The geometry transformer $\mathcal{G}$ comprises $L$ cascaded blocks, each executing three sequential attention operations:
\begin{itemize}
    \item \textbf{Intra-View Local Attention} facilitates fine-grained token interactions within individual images.
    \item \textbf{Cross-View Spatial Attention} enables spatial reasoning across the $V$ views of the current frame.
    \item \textbf{Temporal Causal Attention} performs temporal aggregation between the current frame and the historical cache.
\end{itemize}
In the Temporal Causal Attention module, the current frame's tokens serve as the query, while the cached features of past $W$ historical frames act as the keys and values. 
Crucially, to support infinite-length streaming inference, we discard conventional absolute temporal positional encoding and adopt MRoPE-I~\cite{mrope} for relative temporal positional encoding instead. 
This design ensures that the cached historical features remain invariant over time, allowing them to be reused directly for future interactions without recomputation.
After processing the current frame, the cache is updated following a First-In-First-Out (FIFO) principle. 
The earliest frame's features are discarded, and the current intermediate features $\mathbf{\hat{G}}_t$ of each transformer layer are pushed into the cache:
\begin{equation}
    \mathbf{C}_{t-W+1:t} = \text{FIFO}(\mathbf{C}_{t-W:t-1}, \mathbf{\hat{G}}_t),
\end{equation}
This sliding-window mechanism ensures that the model avoids redundant computation for historical frames, maintaining a constant and fast inference speed.

\textbf{Prediction Heads}. 
We employ three specialized heads to decode the output representations of the geometry transformer.
The visual tokens are processed by a DPT head~\cite{dpt} $\mathcal{H}^{vis}$ to recover the dense 3D pointmaps $\mathbf{P}_t$. 
For pose and trajectory prediction, we first aggregate the corresponding tokens across all views.
Inspired by DiffusionDrive~\cite{diffusiondrive}, we then employ two anchor-based diffusion heads $\mathcal{H}^{\text{pose}}$ and $\mathcal{H}^{\text{traj}}$ to model the prior distributions of the ego-pose and ego-trajectory, respectively.
By leveraging a truncated diffusion strategy, these heads robustly decode the relative pose $\mathbf{E}_t$ and the future driving trajectory $\mathbf{A}_t$.

\section{Experiments}
\label{sec: experiments}

\subsection{Dataset}
We conduct our training and evaluation on a large-scale mixed driving dataset comprising multi-view video sequences sampled at 2 Hz from five sources: nuScenes (700 train / 150 test scenes), OpenScene (19,736 train / 2,026 test scenes), Waymo (798 train / 202 test scenes), KITTI (138 train / 13 test scenes), and DDAD (150 train / 50 test scenes). To obtain high-quality geometric supervision, we follow DVGT~\cite{dvgt} and employ the depth foundation model MoGe-2~\cite{moge-2} to infer dense depth maps, followed by a threshold-filtering operation to ensure data quality. Finally, to comprehensively assess planning performance, we evaluate our model on the closed-loop NAVSIM v1 and v2 benchmarks~\cite{navsim-v1, navsim-v2} (curated subsets of OpenScene), as well as the standard open-loop nuScenes benchmark.

\begin{table*}[t] \small
    \caption{\textbf{Quantitative 3D geometry reconstruction results on OpenScene~\cite{openscene}.} * denotes performing post-alignment with sparse LiDAR to recover the metric scale.}
    \vspace{-3mm}
    \label{tab:openscene_geometry}
    \centering
    \setlength{\tabcolsep}{3pt}
    \renewcommand{\tabcolsep}{4pt}     %
    \scalebox{0.9}{
    \begin{tabular}{cc|c|c|c|c|c|c}
        \toprule
        
         {\textbf{Method}} & Paradigm & {Acc $\downarrow$} & {Comp $\downarrow$} & {Abs Rel $\downarrow$} & {$\delta$\textless{}$1.25\uparrow$} & {AUC $\uparrow$} & {Time} \\ 
        \midrule

        VGGT*~\cite{vggt} & Full-Seq. & 1.705 & 1.711 & 0.280 & 0.669 & \underline{74.4} & $\sim$5.31s \\
        
        MapAnything~\cite{mapanything} & Full-Seq. & 3.269 & 4.214 & 0.476 & 0.253 & 69.5 & $\sim$2.28s \\

        DVGT~\cite{dvgt} & Full-Seq. & \textbf{0.412} & \underline{0.491} & \underline{0.048} & \underline{0.971} & \textbf{76.6} & $\sim$1.88s \\

        \midrule
        
        CUT3R*~\cite{cut3r} & Streaming & 1.858 & 2.245 & 0.275 & 0.596 & 36.9 & $\sim$\underline{0.35}s \\

        StreamVGGT*~\cite{streamvggt} & Streaming & 2.209 & 2.060 & 0.303 & 0.620 & 74.1 & $\sim$1.94s \\

        Driv3R*~\cite{driv3r} & Streaming & 0.884 & 1.693 & 0.188 & 0.740 & - & $\sim$0.56s \\

        \model & Streaming & \underline{0.440} & \textbf{0.450} & \textbf{0.040} & \textbf{0.977} & 70.3 & $\sim$\textbf{0.27}s \\
        
        \bottomrule
    \end{tabular}
    }
    \vspace{-3mm}
\end{table*}

\begin{table*}[t] \small
    \caption{\textbf{Quantitative 3D geometry reconstruction results on nuScenes~\cite{nuscenes}.}}
    \vspace{-3mm}
    \label{tab:nuscenes_geometry}
    \centering
    \setlength{\tabcolsep}{3pt}
    \renewcommand{\tabcolsep}{6.66pt}     %
    \scalebox{0.9}{
    \begin{tabular}{cc|c|c|c|c|c}
        \toprule
        
         {\textbf{Method}} & Paradigm & {Acc $\downarrow$} & {Comp $\downarrow$} & {Abs Rel $\downarrow$} & {$\delta$\textless{}$1.25\uparrow$} & AUC $\uparrow$ \\ 
        \midrule

        VGGT*~\cite{vggt} & Full-Seq. & 1.944 & 2.071 & 0.348 & 0.573 & 83.9 \\
        
        MapAnything~\cite{mapanything} & Full-Seq. & 4.447 & 4.860 & 0.562 & 0.271 & 84.9 \\

        DVGT~\cite{dvgt} & Full-Seq. & \textbf{0.469} & \textbf{0.508} & \underline{0.067} & \underline{0.955} & \textbf{86.4} \\

        \midrule
        
        CUT3R*~\cite{cut3r} & Streaming & 2.055 & 2.611 & 0.330 &  0.547 & 44.5 \\ 

        StreamVGGT*~\cite{streamvggt} & Streaming & 2.414 & 2.284 & 0.375 & 0.563 & \underline{86.0} \\

        Driv3R*~\cite{driv3r} & Streaming & \underline{0.742} & 1.345 & 0.189 &  0.721 & -- \\

        \model & Streaming & 0.775 & \underline{0.792} & \textbf{0.055} & \textbf{0.965} & 84.5 \\
        
        \bottomrule
    \end{tabular}
    }
    \vspace{-7mm}
\end{table*}

\subsection{Implementation Details}
\textbf{Architecture.}
Following DVGT~\cite{dvgt}, we utilize a ViT-L pretrained by DINOv3~\cite{dinov3} as the image encoder. 
The subsequent geometry transformer is composed of $L = 24$ blocks, where each block consists of an intra-view local attention
layer, a cross-view spatial attention layer, and a temporal causal attention layer. 
All attention layers operate with a feature dimension of 1024 and 16 heads.
For the anchor-based diffusion heads, we build upon DiffusionDrive~\cite{diffusiondrive} with minor architectural modifications to predict ego-poses and future trajectories. 
To enhance trajectory planning, we integrate ego status, comprising velocity, acceleration, and driving command, into the trajectory token via an MLP.

\textbf{Training.}
We train our general \model on the mixed dataset using a two-stage paradigm. 
In the first stage, we conduct geometry reconstruction pre-training without enabling the streaming mechanism. 
In the second stage, we conduct Vision-Geometry-Action training by introducing trajectory planning supervision and incorporating the streaming mechanism. 
During both stages, we randomly sample sequences of 2 to 8 views and 2 to 24 frames per scene for training.
To further enhance closed-loop planning, we also perform fine-tuning on the NAVSIM to yield the specialized \textbf{DVGT-2-NAVSIM}, which is trained on sequences with fixed 8 views and 4 frames.
The entire training process takes approximately ten days on 64 H20 GPUs.
  
\begin{table*}[t] \small
    \caption{\textbf{Quantitative 3D geometry reconstruction results on Waymo~\cite{waymo}.}}
    \vspace{-3mm}
    \label{tab:waymo_geometry}
    \centering
    \setlength{\tabcolsep}{3pt}
    \renewcommand{\tabcolsep}{6.66pt}     %
    \scalebox{0.9}{
    \begin{tabular}{cc|c|c|c|c|c}
        \toprule
        
         {\textbf{Method}} & Paradigm & {Acc $\downarrow$} & {Comp $\downarrow$} & {Abs Rel $\downarrow$} & {$\delta$\textless{}$1.25\uparrow$} & AUC $\uparrow$ \\ 
        \midrule

        VGGT*~\cite{vggt} & Full-Seq. & 3.202 & 3.390 & 0.333 & 0.567 & 84.3 \\
        
        MapAnything~\cite{mapanything} & Full-Seq. & 9.903 & 8.365 & 0.492 & 0.217 & 82.3 \\ 

        DVGT~\cite{dvgt} & Full-Seq. & 1.752 & 2.263 & \underline{0.102} & \underline{0.922} & \textbf{86.0} \\

        \midrule
        
        CUT3R*~\cite{cut3r} & Streaming & 3.369 & 4.192 & 0.290 &  0.563 & 50.9 \\

        StreamVGGT*~\cite{streamvggt} & Streaming & 3.207 & 2.960 & 0.302 & 0.604 & 85.6 \\

        Driv3R*~\cite{driv3r} & Streaming & \textbf{0.800} & \textbf{1.311} & 0.168 &  0.770 & -- \\

        \model & Streaming & \underline{1.238} & \underline{1.367} & \textbf{0.073} & \textbf{0.949} & \underline{85.9} \\
        
        \bottomrule
    \end{tabular}
    }
    \vspace{-3mm}
\end{table*}

\begin{table*}[t] \small
    \caption{\textbf{Quantitative 3D geometry reconstruction results on DDAD~\cite{ddad}.}}
    \vspace{-3mm}
    \label{tab:ddad_geometry}
    \centering
    \setlength{\tabcolsep}{3pt}
    \renewcommand{\tabcolsep}{6.66pt}     %
    \scalebox{0.9}{
    \begin{tabular}{cc|c|c|c|c|c}
        \toprule
        
         {\textbf{Method}} & Paradigm & {Acc $\downarrow$} & {Comp $\downarrow$} & {Abs Rel $\downarrow$} & {$\delta$\textless{}$1.25\uparrow$} & AUC $\uparrow$ \\ 
        \midrule

        VGGT*~\cite{vggt} & Full-Seq. & 2.322 & 2.879 & 0.798 & 0.395 & 86.3 \\
        
        MapAnything~\cite{mapanything} & Full-Seq. & 7.442 & 7.928 & 1.836 & 0.207 & 87.4 \\

        DVGT~\cite{dvgt} & Full-Seq. & \textbf{0.751} & \textbf{1.017} & \underline{0.145} & \underline{0.848} & \textbf{95.3} \\

        \midrule
        
        CUT3R*~\cite{cut3r} & Streaming & 2.827 & 4.724 & 0.875 & 0.317 & 48.6 \\ 

        StreamVGGT*~\cite{streamvggt} & Streaming & 2.655 & 2.731 & 0.810 & 0.419 & 91.9 \\

        Driv3R*~\cite{driv3r} & Streaming & \underline{0.950} & \underline{1.259} & 0.185 & 0.740 & -- \\

        \model & Streaming & 1.770 & 1.837 & \textbf{0.093} & \textbf{0.919} & \underline{92.5} \\
        
        \bottomrule
    \end{tabular}
    }
    \vspace{-7mm}
\end{table*}

\subsection{Evaluation Metrics}
\textbf{Geometry Reconstruction}.
Following DVGT~\cite{dvgt}, we assess 3D pointmap quality using \textit{Accuracy} (proximity of predicted points to the ground truth) and \textit{Completeness} (coverage of the ground truth). 
We further evaluate the ray depth of the pointmaps using \textit{Abs Rel} (Absolute Relative error) and \textit{$\delta < 1.25$} (threshold accuracy). 
For ego-pose estimation, we report the \textit{AUC} to measure the Area Under the Curve of the relative pose error.

\textbf{Planning}.
We conduct both open-loop and closed-loop evaluations.
For open-loop planning on nuScenes, we measure the L2 displacement error and collision rate over a 3-second future horizon.
For closed-loop planning, we utilize the NAVSIM v1 and v2 benchmarks. 
NAVSIM v1 simulates a 4-second non-reactive environment at 10 Hz, scoring agents via the Predictive Driver Model Score (PDMS), which aggregates fundamental safety, comfort, and progress metrics. 
NAVSIM v2 enhances simulation realism with reactive traffic and introduces the Extended PDMS (EPDMS), which incorporates additional criteria such as traffic rule compliance and extended comfort.

\begin{table*}[t]
\caption{\textbf{Closed-loop planning results on NAVSIM v1 \texttt{navtest} split}. $\dagger$ denotes using reinforcement learning to boost planning scores. Future states represent world-modeling-based methods. C and L denote camera and LiDAR, respectively.}
\vspace{-4mm}
\label{tab:navsim}
\centering
\resizebox{1.0\textwidth}{!}{
\begin{tabular}{l|cc|cccccc}
    \toprule
    Method & Input & Aux. Sup. & NC $\uparrow$ &DAC $\uparrow$ & TTC $\uparrow$& Comf. $\uparrow$ & EP $\uparrow$ & \cellcolor{gray!30}PDMS $\uparrow$  \\
    \midrule
    PARA-Drive~\cite{para-drive} & C & Map \& Mot. \& Occ & 97.9 & 92.4 & 93.0 & 99.8 & 79.3 & \cellcolor{gray!30}84.0 \\
    VADv2~\cite{vadv2} & C & Map \& Mot. \& Traffic & 97.2 & 89.1 & 91.6 & 100 & 76.0 & \cellcolor{gray!30}80.9 \\
    UniAD~\cite{uniad} & C & Map \& Box \& Mot. \& Occ & 97.8 & 91.9 & 92.9 & 100 & 78.8 & \cellcolor{gray!30}83.4 \\
    Transfuser~\cite{transfuser} & C \& L & Map \& Box & 97.7 & 92.8 & 92.8 & 100 & 79.2 & \cellcolor{gray!30}84.0 \\
    Hydra-MDP~\cite{hydra-mdp} & C \& L & Map \& Box & 98.3 & 96.0 & 94.6 & 100 & 78.7 & \cellcolor{gray!30}86.5 \\
    GoalFlow~\cite{goalflow} & C \& L & Map \& Box & 98.3 & 93.8 & 94.3 & 100 & 79.8 & \cellcolor{gray!30}85.7 \\
    ARTEMIS~\cite{artemis} & C \& L & Map \& Box & 98.3 & 95.1 & 94.3 & 100 & 81.4 & \cellcolor{gray!30}87.0 \\
    DiffusionDrive~\cite{diffusiondrive} & C \& L & Map \& Box & 98.2 & 96.2 & 94.7 & 100 & 82.2 & \cellcolor{gray!30}88.1 \\
    WoTE~\cite{wote} & C \& L & Map \& Box & 98.5 & 96.8 & 94.9 & 99.9 & 81.9 & \cellcolor{gray!30}88.3 \\
    DriveSuprim~\cite{drivesuprim} & C \& L & Map \& Box & 97.8 & 97.3 & 93.6 & 100 & 86.7 & \cellcolor{gray!30}89.9 \\ 
    \midrule
    AutoVLA~\cite{autovla} & C & Language & 96.9 & 92.4 & 88.1 & 99.9 & 75.8 & \cellcolor{gray!30}80.5 \\
    AdaThinkDrive\cite{luo2025adathinkdrive} & C & Language &98.5 &94.4 &94.9 &100 &79.9 &\cellcolor{gray!30}86.2 \\
    ReCogDrive~\cite{recogdrive} & C & Language & 98.3 & 95.1 & 94.3 & 100 & 81.1 & \cellcolor{gray!30}86.8 \\
    DriveVLA-W0~\cite{drivevla-w0} & C & Future States & 98.7 & 99.1 & 95.3 & 99.3 & 83.3 & \cellcolor{gray!30}\textbf{90.2} \\
    \midrule
    AutoVLA$^\dagger$~\cite{autovla} & C & Language \& RL & 98.4 & 95.6 & 98.0 & 99.9 & 85.9 & \cellcolor{gray!30}89.1 \\
    ReCogDrive$^\dagger$~\cite{recogdrive} & C & Language \& RL & 98.2 & 97.8 & 95.2 & 99.8 & 83.5 & \cellcolor{gray!30}89.6 \\
    \midrule
    \model & C & Dense Geometry & 97.8 & 97.2 & 93.9 & 100 & 83.4 & \cellcolor{gray!30}{88.6}\\
    \textbf{DVGT-2-NAVSIM} & C & Dense Geometry & 98.7 & 97.9 & 95.8 & 100 & 84.3 & \cellcolor{gray!30}\textbf{90.3}\\
    \bottomrule
\end{tabular}%
}
\vspace{-4mm}
\end{table*}

\begin{table*}[t]
\caption{\textbf{Closed-loop planning results on NAVSIM v2 \texttt{navtest} split}.}
\vspace{-4mm}
\label{tab:navsimv2}
\centering
\resizebox{1.0\textwidth}{!}{
\begin{tabular}{l|cccccccccc}
    \toprule
    Method & NC $\uparrow$ & DAC $\uparrow$ & DDC $\uparrow$ & TL $\uparrow$ & EP $\uparrow$ & TTC $\uparrow$ & LK $\uparrow$ & HC $\uparrow$ & EC $\uparrow$ & \cellcolor{gray!30}EPDMS $\uparrow$ \\
    \midrule
    Ego Status MLP & 93.1 & 77.9 & 92.7 & 99.6 & 86.0 & 91.5 & 89.4 & 98.3 & 85.4 & \cellcolor{gray!30}64.0 \\
    Transfuser~\cite{transfuser} & 96.9 & 89.9 & 97.8 & 99.7 & 87.1 & 95.4 & 92.7 & 98.3 & 87.2 & \cellcolor{gray!30}76.7 \\
    Hydra-MDP++~\cite{hydra-mdp++} & 97.2 & 97.5 & 99.4 & 99.6 & 83.1 & 96.5 & 94.4 & 98.2 & 70.9 & \cellcolor{gray!30}81.4 \\
    DriveSuprim~\cite{drivesuprim} & 97.5 & 96.5 & 99.4 & 99.6 & 88.4 & 96.6 & 95.5 & 98.3 & 77.0 & \cellcolor{gray!30}83.1 \\
    ARTEMIS~\cite{artemis} & 98.3 & 95.1 & 98.6 & 99.8 & 81.5 & 97.4 & 96.5 & 98.3 & - & \cellcolor{gray!30}83.1 \\
    DiffusionDrive~\cite{diffusiondrive} & 98.2 & 95.9 & 99.4 & 99.8 & 87.5 & 97.3 & 96.8 & 98.3 & 87.7 & \cellcolor{gray!30}84.5 \\
    DriveVLA-W0~\cite{drivevla-w0} & 98.5 & 99.1 & 98.0 & 99.7 & 86.4 & 98.1 & 93.2 & 97.9 & 58.9 & \cellcolor{gray!30}86.1 \\
    \midrule
    \model & 97.8 & 97.2 & 99.6 & 99.9 & 88.4 & 97.3 & 98.1 & 98.2 & 83.2 & \cellcolor{gray!30}\underline{88.9} \\
    \textbf{DVGT-2-NAVSIM} & 98.7 & 97.9 & 99.7 & 99.9 & 87.9 & 98.0 & 98.2 & 98.2 & 77.0 & \cellcolor{gray!30}\textbf{89.6} \\
    \bottomrule
\end{tabular}%
}
\vspace{-7mm}
\label{tab:navsimv2}
\end{table*}

\subsection{Geometry Reconstruction}

\textbf{Inference Settings.}
We evaluate geometry reconstruction through online, frame-by-frame prediction on 16-frame multi-view sequences. 
All reported metrics, including global point reconstruction, local ray depth estimation, and global ego-pose prediction, are averaged over 16 frames of the sequences. 
For non-streaming methods (e.g., VGGT, MapAnything, and DVGT), we implement a streaming inference paradigm, where the inference is performed incrementally by adding one frame per step.
For our model, we maintain a history cache with a length of $W=4$ for efficient geometry reconstruction and trajectory planning.
Notably, while existing methods predict the global pointmaps and ego-poses in the first frame's coordinate system, our \model predicts local geometry and relative ego-poses at each frame. 
For a fair comparison, we iteratively transform our predicted local pointmaps and ego-poses into the global coordinate system by accumulating the predicted relative ego-poses before metric evaluation.

\textbf{Ray Depth Estimation.}
Ray depth is defined as the distance from a 3D point to the current ego center, which serves as a critical indicator of local geometric accuracy. 
Since our model natively predicts local pointmaps rather than global ones, it inherently preserves local structural details more effectively. 
As shown in ~\cref{tab:openscene_geometry,tab:nuscenes_geometry,tab:waymo_geometry,tab:ddad_geometry}, our approach achieves the state-of-the-art ray depth performance across multiple datasets, outperforming both the general vision-geometry models~\cite{vggt, streamvggt} and the driving-specific models~\cite{driv3r,dvgt}.

\textbf{Global Point Reconstruction.}
Evaluating our model on global point reconstruction is inherently challenging. 
Our local pointmap predictions require iterative aggregation via the predicted relative ego-poses to construct a global pointmap.
This process inevitably introduces cumulative errors from ego-pose estimation, while other methods directly output global pointmaps.
Despite this structural disadvantage, our model achieves strong global point reconstruction performance comparable to existing SOTA methods, and even surpasses them on specific datasets, as shown in ~\cref{tab:openscene_geometry,tab:waymo_geometry}.

\begin{figure}[!t]
\centering
\includegraphics[width=1.0\textwidth]{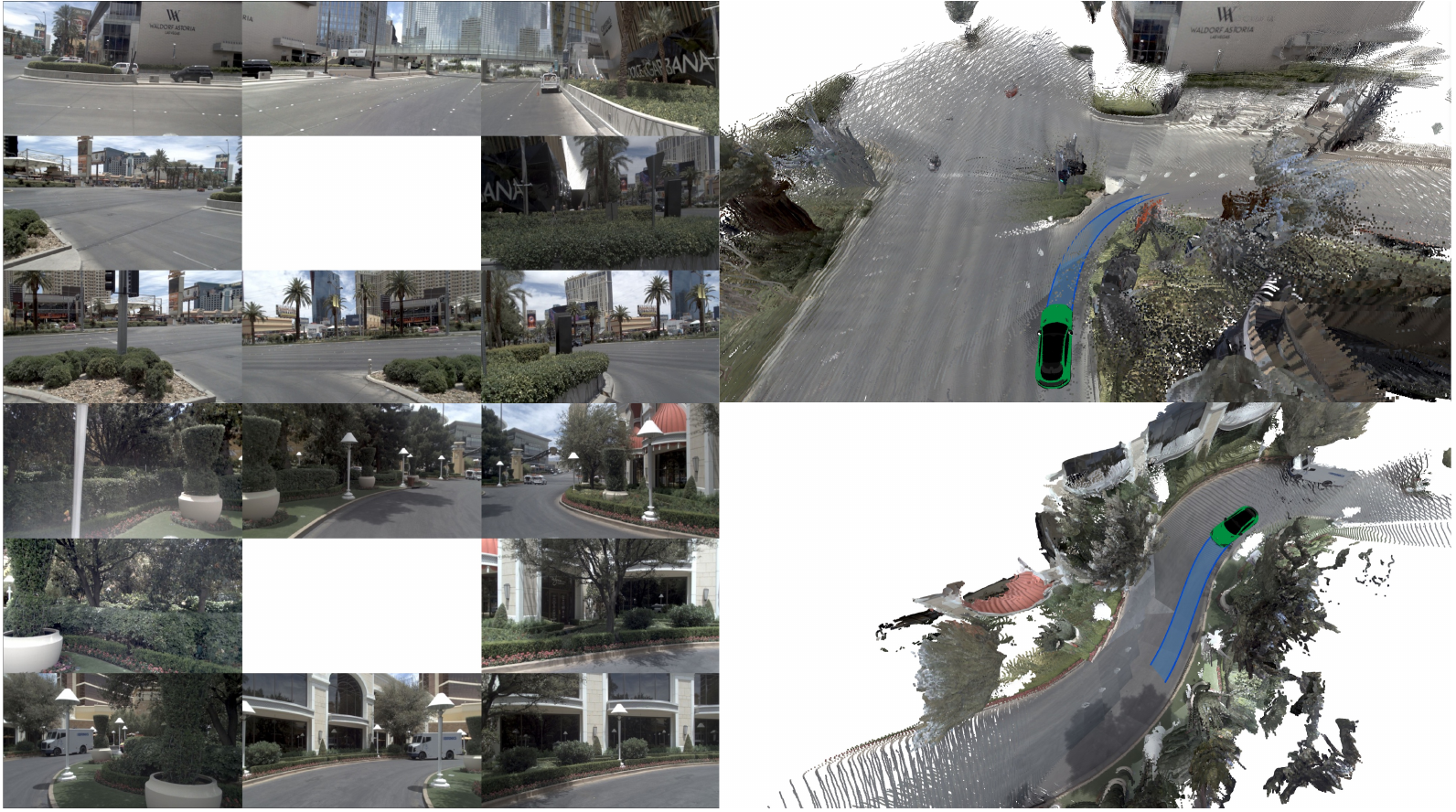}
\vspace{-7mm}
\caption{\textbf{Qualitative visualizations}. These results demonstrate that \model can predict high-fidelity dense scene geometry and perform robust trajectory planning.}
\label{fig:vis}
\vspace{-6mm}
\end{figure}

\textbf{Global Pose Prediction.}
We note that our model is less competitive in global ego-pose estimation. We attribute this to three main factors. 
First, to prioritize inference efficiency, we employ a lightweight pose head with a truncated two-step diffusion strategy. 
In contrast, baseline models like VGGT and DVGT rely on a heavier four-step residual reasoning process, which yields higher accuracy but degrades inference efficiency. 
Second, similar to the global pointmap evaluation, deriving global ego-poses by accumulating relative poses inevitably introduces trajectory drift over time. 
Finally, our sliding window streaming strategy restricts the temporal context to a fixed historical window for online inference. 
While this enables efficient per-frame processing, it lacks the global context utilized by baseline methods that access the entire sequence simultaneously, thereby limiting long-term global ego-pose consistency.

\textbf{Inference Efficiency.}
We compare the online inference efficiency of different methods on the OpenScene dataset using 16-frame, 8-view sequences. 
Compared to batch-processing methods and full-history streaming methods, our proposed sliding window streaming strategy significantly accelerates online inference. 
As detailed in ~\cref{tab:openscene_geometry}, with a fixed window size of 4, our method efficiently processes the 16-frame sequence with an average latency of only 0.27s per frame. 
This outperforms prior methods while maintaining robust reconstruction capabilities, demonstrating its critical value for real-time autonomous driving.

\begin{table*}[t]
\caption{\textbf{Open-looped planning results on nuScenes~\cite{nuscenes}.} 
$\dagger$ denotes the results computed with an average of previous frames as adopted in VAD~\cite{vad}.
Aux. Sup. represents auxiliary supervision.
Avg. computes the average result of 1s, 2s, and 3s.
}
\vspace{-4mm}
\centering
\setlength{\tabcolsep}{0.005\linewidth}
\scriptsize %
\resizebox{1.0\textwidth}{!}{
\begin{tabular}{l|cc|cccc|cccc}
\toprule
\multirow{2}{*}{Method} & \multirow{2}{*}{Input} & \multirow{2}{*}{Aux. Sup.} &
\multicolumn{4}{c|}{L2 (m) $\downarrow$} & 
\multicolumn{4}{c}{Collision Rate (\%) $\downarrow$} \\
&& & 1s & 2s & 3s & \cellcolor{gray!30}Avg. & 1s & 2s & 3s & \cellcolor{gray!30}Avg. \\
\midrule
IL~\cite{IL} & LiDAR & None  & 0.44 & 1.15 & 2.47 & \cellcolor{gray!30}1.35 & 0.08 & 0.27 & 1.95 & \cellcolor{gray!30}0.77 \\
NMP~\cite{NMP} & LiDAR & Box \& Motion & 0.53 & 1.25 & 2.67 & \cellcolor{gray!30}1.48 & {0.04} & \underline{0.12} & 0.87 & \cellcolor{gray!30}0.34 \\
FF~\cite{FF} & LiDAR & Freespace  & 0.55 & 1.20 & 2.54 & \cellcolor{gray!30}1.43 & 0.06 & 0.17 & 1.07 & \cellcolor{gray!30}0.43 \\
EO~\cite{EO} & LiDAR & Freespace  & 0.67 & 1.36 & 2.78 & \cellcolor{gray!30}1.60 & {0.04} & 0.09 & 0.88 & \cellcolor{gray!30}0.33 \\
\midrule
ST-P3~\cite{hu2022st} & Camera & Map \& Box \& Depth & 1.33 & 2.11 & 2.90 & \cellcolor{gray!30}2.11 & 0.23 & 0.62 & 1.27 & \cellcolor{gray!30}0.71 \\
UniAD~\cite{uniad} & Camera & { Map \& Box \& Mot. \& Occ}  & {0.48} & 0.96 & 1.65 & \cellcolor{gray!30}1.03 & 0.05 & 0.17 & \underline{0.71}& \cellcolor{gray!30}\underline{0.31} \\
VAD-Tiny~\cite{vad}  & Camera & Map \& Box \& Mot.  & 0.60 & 1.23 & 2.06 & \cellcolor{gray!30}1.30 & 0.31 & 0.53 & 1.33 & \cellcolor{gray!30}0.72 \\
VAD-Base~\cite{vad} & Camera & Map \& Box \& Mot. & 0.54 & 1.15 & 1.98 & \cellcolor{gray!30}1.22 & {0.04} & 0.39 & 1.17 & \cellcolor{gray!30}0.53 \\
GenAD~\cite{genad} & Camera & Map \& Box \& Mot. & \underline{0.36} & 0.83& 1.55& \cellcolor{gray!30}0.91& 0.06 & {0.23} & 1.00 & \cellcolor{gray!30}0.43  \\
OccWorld~\cite{occworld} & Camera & Occ & 0.52 & 1.27 & 2.41 & \cellcolor{gray!30}1.40 & 0.12 & 0.40 & 2.08 & \cellcolor{gray!30}0.87  \\
OccNet~\cite{occnet} & Camera & Occ \& Map \& Box  & 1.29 & 2.13 & 2.99 & \cellcolor{gray!30}2.14 & 0.21 & 0.59 & 1.37 & \cellcolor{gray!30}0.72  \\
GaussianAD~\cite{gaussianad} & Camera & Map \& Box \& Mot. \& Occ & 0.40& \textbf{0.64}& \textbf{0.88}& \cellcolor{gray!30}\textbf{0.64}& 0.09 & 0.38& 0.81& \cellcolor{gray!30}0.42\\
\midrule
OmniDrive~\cite{omnidrive} & Camera & Map \& Box \& Language & 0.40 & 0.80 & \underline{1.32} & \cellcolor{gray!30}0.84 & {0.04} & 0.46 & 2.32 & \cellcolor{gray!30}0.94  \\
Doe-1~\cite{doe-1} & Camera & Language & 0.50 & 1.18 & 2.11 & \cellcolor{gray!30}1.26 & {0.04} & 0.37 & 1.19 & \cellcolor{gray!30}0.53  \\
UniUGP~\cite{lu2025uniugp} & Camera & Language & 0.58 & 1.14 & 1.95 & \cellcolor{gray!30}1.23 & \underline{0.01} & 0.19 & 0.81 & \cellcolor{gray!30}0.33  \\
\midrule
\textbf{\model}& Camera & Dense Geometry & \textbf{0.25} & \underline{0.67} & {1.43} & \cellcolor{gray!30} \underline{0.78} & \textbf{0.00} & \textbf{0.07} & \textbf{0.50} & \cellcolor{gray!30}\textbf{0.19} \\
\midrule
\color{gray}VAD-Tiny$^\dagger$~\cite{vad}  & \color{gray}Camera & \color{gray}Map \& Box \& Mot.  & \color{gray}0.46 & \color{gray}0.76 & \color{gray}1.12 & \color{gray}\cellcolor{gray!30}0.78 & \color{gray}0.21 & \color{gray}0.35 & \color{gray}0.58 & \color{gray}\cellcolor{gray!30}0.38 \\
\color{gray}VAD-Base$^\dagger$~\cite{vad} & \color{gray}Camera & \color{gray}Map \& Box \& Mot. & \color{gray}0.41 & \color{gray}0.70 & \color{gray}1.05 & \color{gray}\cellcolor{gray!30}0.72 & \color{gray}\underline{0.07} & \color{gray}0.17 & \color{gray}\underline{0.41} & \color{gray}\cellcolor{gray!30}\underline{0.22} \\
\color{gray}{OccWorld-D}$^\dagger$~\cite{occworld} & \color{gray}Camera & \color{gray}Occ & \color{gray}0.39 & \color{gray}0.73 & \color{gray}1.18 & \color{gray}\cellcolor{gray!30} 0.77 & \color{gray}0.11 & \color{gray}0.19 & \color{gray}0.67 & \color{gray}\cellcolor{gray!30} 0.32  \\
\color{gray}GenAD$^\dagger$~\cite{genad} & \color{gray}Camera & \color{gray}Map \& Box \& Mot. & \color{gray}\underline{0.28} & \color{gray}0.49& \color{gray}{0.78}& \color{gray}\cellcolor{gray!30}{0.52}& \color{gray}{0.08} & \color{gray}\textbf{0.14} & \color{gray}\textbf{0.34} & \color{gray}\cellcolor{gray!30}\textbf{0.19}  \\
\color{gray}GaussianAD$^\dagger$\cite{gaussianad} & \color{gray}Camera & \color{gray}Map \& \color{gray}Box \& \color{gray}Occ   & \color{gray}0.34& \color{gray}\underline{0.47}& \color{gray}\textbf{0.60}& \cellcolor{gray!30}\color{gray}\underline{0.47}& \color{gray}0.49& \color{gray}0.49& \color{gray}0.51& \cellcolor{gray!30}\color{gray}0.50\\
\midrule
\color{gray}\textbf{\model}$^\dagger$ & \color{gray}Camera & \color{gray}Dense Geometry & \color{gray}\textbf{0.20}& \color{gray}\textbf{0.37}& \color{gray}\underline{0.66}& \cellcolor{gray!30}\color{gray}\textbf{0.41}& \color{gray}\textbf{0.04}& \color{gray}\textbf{0.14}& \color{gray}0.47& \cellcolor{gray!30}\color{gray}\underline{0.22}\\
\bottomrule
\end{tabular} }
\label{tab:sota-plan}
\vspace{-7mm}
\end{table*}

\subsection{Planning}
\textbf{Closed-loop Planning on NAVSIM v1~\cite{navsim-v1} and v2~\cite{navsim-v2}}.
Trained on a large mixture of datasets, our foundation VGA model \model performs robust planning based on high-fidelity dense geometry reconstruction. 
As shown in ~\cref{tab:navsim}, \model achieves an 88.6 PDMS on NAVSIM v1, comparable to SOTA end-to-end and VLA models. 
It also yields an EPDMS of 88.9 on NAVSIM v2 (~\cref{tab:navsimv2}), outperforming all SOTA methods. 
We also fine-tune this foundation model on NAVSIM to produce \textbf{DVGT-2-NAVSIM}, which establishes new SOTA on both benchmarks. 
We show two examples of geometry reconstruction and planning by \model in~\cref{fig:vis}.
These results validate dense geometry as a robust foundation for safe planning.
Unlike traditional end-to-end models that rely on multi-modal inputs and sparse perceptual supervision, or VLA models that require language labels and complicated RL fine-tuning, our VGA paradigm achieves safe, end-to-end driving using only annotation-efficient geometric supervision.

\textbf{Open-loop Planning on nuScenes}. 
As shown in ~\cref{tab:sota-plan}, \model achieves L2 error metrics comparable to SOTA models on nuScenes. 
More importantly, \model yields a significantly lower collision rate than models explicitly trained with high-level semantic labels (which directly define the collision metrics). 
This highlights that our model inherently learns the comprehensive 3D structure and physical interactions between the ego-vehicle and the environment, enabling robust planning without relying on sparse perceptual annotations.

\subsection{Inference Efficiency Comparison}
We compare the online inference efficiency of different methods, focusing on per-frame memory cost and latency. All models perform frame-by-frame inference on identical multi-frame, 8-view sequences. It is worth noting that our model simultaneously performs geometry reconstruction and trajectory planning, whereas the compared models focus solely on geometry reconstruction. 

\begin{figure}[t]
  \centering
  \begin{subfigure}[b]{0.50\textwidth}
    \centering
    \includegraphics[width=\linewidth]{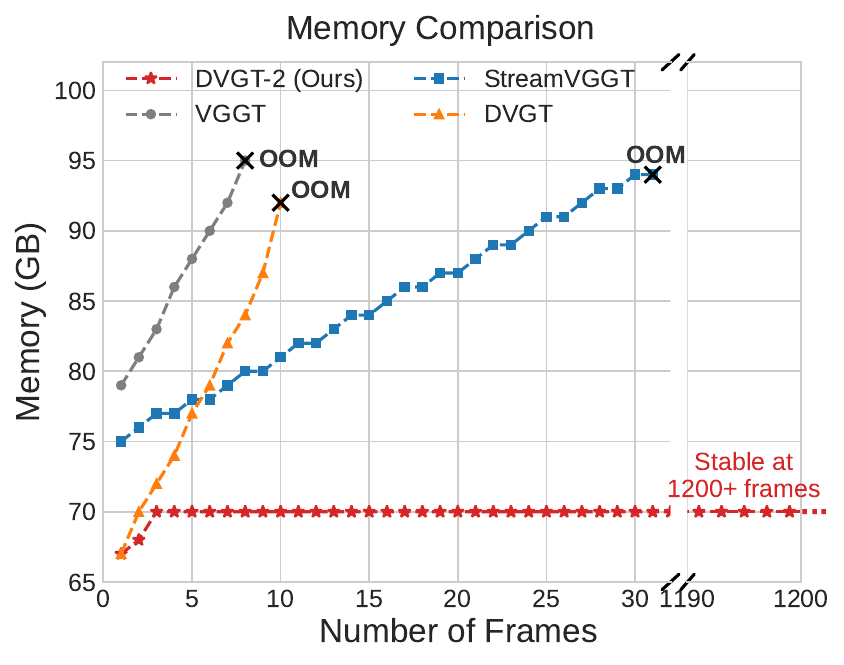}
    \vspace{-4mm}
    \caption{Memory Comparison.}
    \label{fig:memory}
  \end{subfigure}
  \hfill
  \begin{subfigure}[b]{0.48\textwidth} %
    \centering
    \includegraphics[width=\linewidth]{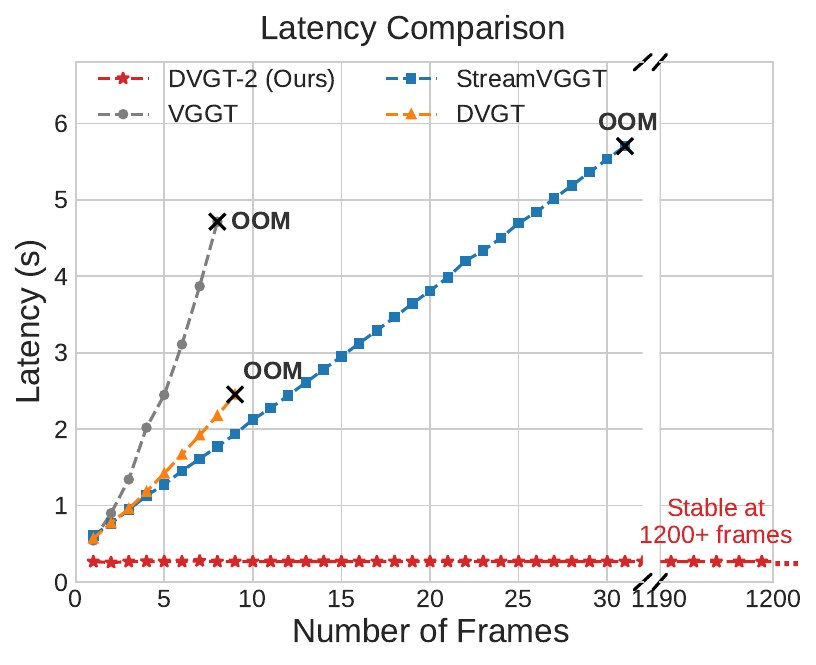}
    \vspace{-4mm}
    \caption{Latency Comparison.}
    \label{fig:latency}
  \end{subfigure}
  \vspace{-3mm}
  \caption{\textbf{Efficiency comparison of online inference}. We report the per-frame latency and memory cost of different methods on multi-frame, 8-view sequences.}
  \vspace{-7mm}
\end{figure}

As shown in Fig.~\ref{fig:memory}, VGGT and DVGT encounter out-of-memory (OOM) errors after only about 10 frames. This is because their full-sequence reconstruction incurs a quadratic memory complexity of $O(T^2)$ with respect to the sequence length $T$. StreamVGGT's memory grows more slowly but still hits OOM at around 30 frames due to its full-history streaming strategy ($O(T)$ complexity). In contrast, \model maintains a constant memory cost. By employing a sliding window streaming strategy with a fixed-size historical cache, our model achieves an $O(1)$ memory cost, enabling infinite-length online inference for driving scenes.

The inference latency comparison is shown in Fig.~\ref{fig:latency}. Due to the $O(T^2)$ complexity, the latency of VGGT and DVGT surges rapidly, taking several seconds to process just a few frames. StreamVGGT's latency scales linearly ($O(T)$ complexity), reaching 5–6 seconds at 30 frames, which is still prohibitive for real-time driving. However, \model achieves a stable latency of around 260ms per frame, even across hundreds of frames,  confirming that our model is well-suited for real-time, infinite-length autonomous driving.

\section{Conclusion}
\label{sec: conclusion}
In this paper, we introduce \model, a streaming driving visual geometry transformer that pioneers the Vision-Geometry-Action (VGA) paradigm for end-to-end autonomous driving.
By reconstructing dense 3D geometry as the foundation representation, \model provides comprehensive spatial and temporal cues for robust trajectory planning.
To overcome the computational bottleneck of traditional multi-frame processing, we propose a sliding-window streaming strategy with temporal causal attention and feature caching, enabling efficient, on-the-fly joint prediction of geometry and trajectories. 
Extensive experiments demonstrate that \model achieves strong geometry reconstruction performance with significantly reduced latency on diverse datasets. 
Moreover, it exhibits strong planning capabilities across open-loop and closed-loop benchmarks. 
We hope this work paves the way for more efficient, geometry-aware driving systems.

\bibliographystyle{splncs04}
\bibliography{main}

\appendix

\setcounter{figure}{0}
\setcounter{table}{0}
\renewcommand{\thefigure}{A.\arabic{figure}}
\renewcommand{\thetable}{A.\arabic{table}}

\clearpage

\vspace{-3mm}
\begin{center}
    \centering
    \includegraphics[width=1.0\linewidth]{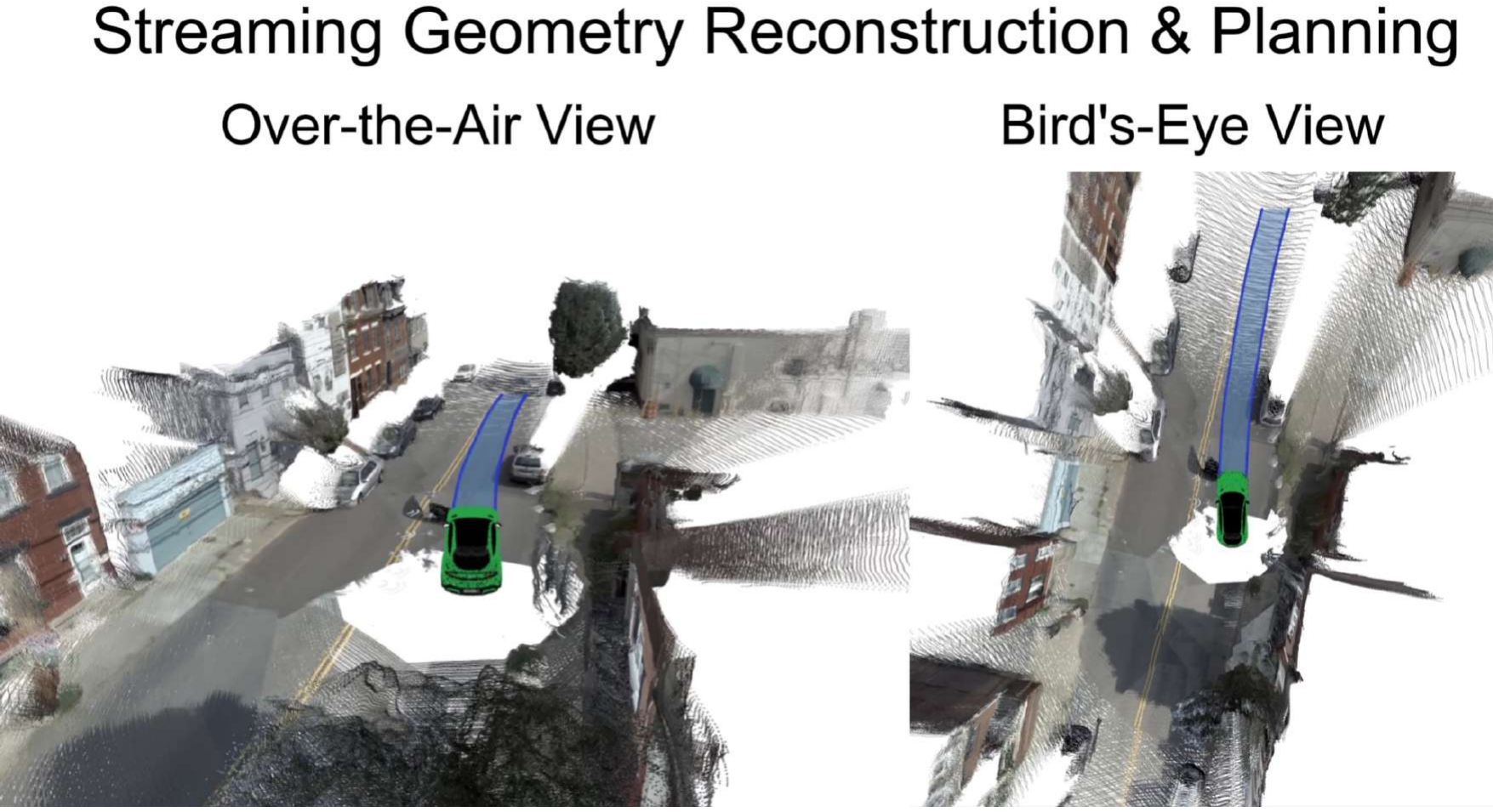}
    \vspace{-7mm}
    \captionof{figure}{\textbf{Video demonstration of \model's geometry reconstruction and trajectory planning based on online multi-view inputs on the validation set.}  
}
\label{fig:demo}
\end{center}%
\vspace{-5mm}

\section{Additional Experiments}

\subsection{Ablation on Window size}
\cref{tab:window_size} details the impact of window size on streaming inference. Increasing the window size from 2 to 6 improves the global point accuracy (Acc) by enlarging the temporal receptive field, which helps model inter-frame relations and global structures. However, a larger window size of 8 slightly degrades the Acc metric. This occurs because the accumulated error of predicted relative ego-poses during the local-to-global point transformation outweighs the benefits of a broader temporal context. Conversely, the ray depth prediction (Abs Rel) remains constant across all settings, indicating that the window size exclusively affects inter-frame and global geometry modeling, rather than local geometry accuracy.

\subsection{Planning on All Datasets}
\cref{tab:planning_L2} presents the 1-second future trajectory prediction performance (L2 error) of \model on the validation sets of five datasets. The model achieves the lowest L2 error of 0.20m on OpenScene, since it makes up over 75\% of the training data. It also maintains competitive performance on NuScenes and Waymo. However, the errors on KITTI and DDAD are notably higher (>2.0m). We attribute this to a significant domain gap in ego-vehicle trajectory distributions, primarily due to much higher driving speeds in these two datasets compared to the others. Since our model employs anchor-based diffusion heads, both the anchor clustering and trajectory distribution modeling are heavily biased towards the dominant OpenScene dataset. Consequently, this leads to sub-optimal performance on KITTI and DDAD in both ego-pose prediction and trajectory planning.

\begin{table}[t]
    \centering
    \begin{minipage}[t]{0.46\linewidth}
        \centering
        \caption{\textbf{Ablation study on the window size of historical cache for streaming inference.}}
        \label{tab:window_size}
        \setlength{\tabcolsep}{1.0mm} 
        \vspace{-3.5mm}
        \begin{tabular}{ccc}
            \toprule
            Window Size & Acc ($\downarrow$) & Abs Rel. ($\downarrow$) \\
            \midrule
            2 & 0.613 & 0.042 \\
            4 & \underline{0.480} & 0.042 \\
            6 & \textbf{0.474} & 0.042 \\
            8 & 0.501 & 0.042 \\
            \bottomrule
        \end{tabular}
        \vspace{-3mm}
    \end{minipage}
    \hfill
    \begin{minipage}[t]{0.46\linewidth}
        \centering
        \caption{\textbf{L2 error of trajectory planning across five datasets.}}
        \label{tab:planning_L2}
        \setlength{\tabcolsep}{3.5mm}
        \vspace{-3.5mm}
        \begin{tabular}{lc}
            \toprule
            Dataset & L2(m) ($\downarrow$) \\
            \midrule
            NuScenes  & 0.56 \\
            OpenScene & 0.20 \\
            Waymo     & 0.78 \\
            KITTI     & 2.12 \\
            DDAD      & 2.00 \\
            \bottomrule
        \end{tabular}
        \vspace{-3mm}
    \end{minipage}
\end{table}

\subsection{Geometry Reconstruction on KITTI}
\begin{table*}[t] \small
    \caption{\textbf{Quantitative 3D geometry reconstruction results on KITTI~\cite{kitti}.} * denotes performing post-alignment with sparse LiDAR to recover the metric scale.}
    \vspace{-3mm}
    \label{tab:kitti_geometry}
    \centering
    \setlength{\tabcolsep}{3pt}
    \renewcommand{\tabcolsep}{6.66pt}     %
    \scalebox{0.9}{
    \begin{tabular}{cc|c|c|c|c|c}
        \toprule
        
         {\textbf{Method}} & Paradigm & {Acc $\downarrow$} & {Comp $\downarrow$} & {Abs Rel $\downarrow$} & {$\delta$\textless{}$1.25\uparrow$} & {AUC@30 $\uparrow$}\\ 
        \midrule

        VGGT*~\cite{vggt} & Full-Seq. & 1.154 &	1.294 &	0.158 &	0.801 &	\textbf{96.91}  \\
        
        MapAnything~\cite{mapanything} & Full-Seq. & 1.807 & \textbf{1.006} & 0.184 & 0.727 & 90.51  \\

        DVGT~\cite{dvgt} & Full-Seq. & \textbf{0.846} & 1.468 & \underline{0.136} & \underline{0.849} & 87.63  \\

        \midrule
        
        CUT3R*~\cite{cut3r} & Streaming & 0.973 & 2.054 & 0.217 & 0.660 & 51.82 \\

        StreamVGGT*~\cite{streamvggt} & Streaming & 3.393 & 2.181 & 0.365 & 0.467 & \underline{95.57} \\

        Driv3R*~\cite{driv3r} & Streaming & \underline{0.864} & \underline{1.083} & 0.164 & 0.784 & -  \\

        \model & Streaming & 1.615 & 3.238 & \textbf{0.087} & \textbf{0.942} & 80.36 \\
        
        \bottomrule
    \end{tabular}
    }
    \vspace{-7mm}
\end{table*}

\cref{tab:kitti_geometry} presents the quantitative geometry reconstruction results on KITTI. \model achieves state-of-the-art performance in ray depth estimation, significantly outperforming other methods in both the Abs Rel and 
$\delta$\textless{}$1.25$ metrics. This superiority stems from our model's inherent design of predicting local pointmaps, which enables effective modeling of local geometry. However, as discussed earlier, the significant domain gap in ego-vehicle trajectories, along with the fact that KITTI accounts for only a small portion of the training data, leads to sub-optimal ego-pose prediction. Furthermore, because the evaluation of global point accuracy relies on these predicted ego-poses to transform local points into global coordinates, accumulated ego-pose errors inevitably degrade the overall performance of global point reconstruction.

\section{Additional Implementation Details}

\textbf{Architecture}.
Building upon the overall architecture in the main paper, our model comprises approximately 1.8 billion parameters. To ensure training stability, we incorporate QKNorm~\cite{qknorm} and LayerScale~\cite{layerscale} (initialized at 0.01) into each attention layer of the geometry transformer. For dense prediction, we follow~\cite{depthanything2} by feeding intermediate tokens from the 4th, 11th, 17th, and 23rd blocks into a DPT~\cite{dpt} head. For ego-pose prediction and trajectory planning, we augment the visual tokens of each view and frame with one pose token and eight trajectory tokens to aggregate global context in the subsequent geometry transformer. Following DiffusionDrive~\cite{diffusiondrive}, we then utilize two anchor-based diffusion heads for decoding, which comprise four self-attention layers for inter-frame interactions and two cross-attention layers for diffusion decoding. We employ 20 anchors for both ego-poses and trajectories, which are pre-computed by clustering the training data.

\textbf{Training}. 
We train the general \model on the mixed dataset for 160K and 80K iterations in the first and second stages, respectively. Subsequently, we finetune the model for 40K iterations on NAVSIM to obtain the specialized \textbf{DVGT-2-NAVSIM}. Across all stages, we optimize using AdamW~\cite{adamw} with a cosine learning rate scheduler, setting the peak learning rate to 1e-4 with an 8K-iteration linear warmup. To ensure training stability and efficiency, we employ gradient norm clipping with a threshold of 1.0, bfloat16 precision, and gradient checkpointing. 
During the first two stages, we train \model on sequences with random views (ranging from 2 to 8) and frames (ranging from 2 to 24) from the mixed dataset. We then finetune \textbf{DVGT-2-NAVSIM} on NAVSIM, where the sequence is fixed to 8 views and 4 frames to align with the standard NAVSIM planning setting. 
For image preprocessing, we first resize the long edge of the input images to 512 pixels while keeping the original aspect ratio. We then center-crop the short edge to a random size between 144 and 320 pixels (ensuring it is divisible by 16). Finally, we apply strong per-frame augmentations—such as color jittering, Gaussian blur, and grayscale conversion—to make the model robust to lighting changes.

\section{Dataset Details}

\begin{table}[t]
    \centering
    \caption{\textbf{Detailed statistics of the datasets used in our experiments.} All temporal statistics are reported at a 2Hz sampling rate.}
    \vspace{-3mm}
    \label{tab:dataset_info}
    \footnotesize 
    \setlength{\tabcolsep}{3pt} %
     \scalebox{1.0}{
    \begin{tabular}{cccccccc}
        \toprule
        \textbf{Dataset} & 
        \makecell{\textbf{Train} \\ \textbf{Scenes}} & 
        \makecell{\textbf{Test} \\ \textbf{Scenes}} & 
        \makecell{\textbf{Min} \\ \textbf{Frames}} & 
        \makecell{\textbf{Max} \\ \textbf{Frames}} & 
        \makecell{\textbf{Avg} \\ \textbf{Frames}} & 
        \makecell{\textbf{Avg Aspect} \\ \textbf{Ratio}} & 
        \makecell{\textbf{Num of} \\ \textbf{Views}} \\ 
        \midrule
        nuScenes  & 700   & 150  & 32 & 41   & 40 & 1.77 & 6 \\
        OpenScene & 19376 & 2026 & 1  & 41   & 38 & 1.77 & 8 \\
        Waymo     & 798   & 202  & 34 & 40   & 40 & 1.77 & 5 \\
        KITTI     & 138   & 13   & 2  & 1033 & 62 & 3.31 & 2 \\
        DDAD      & 150   & 50   & 10 & 20   & 17 & 1.59 & 6 \\ 
        \bottomrule
    \end{tabular}}
    \vspace{-7mm}
\end{table}

\label{sec:dataset_details}
Following DVGT~\cite{dvgt}, we train and evaluate our model on a mixture of five driving datasets: nuScenes, OpenScene, Waymo, KITTI, and c DDAD. \cref{tab:dataset_info} shows their detailed statistics.
All videos are downsampled to 2Hz, and the frame counts in the table are based on this rate.
During training of our general \model, we select datasets for each batch using the following ratio: nuScenes : OpenScene : Waymo : KITTI : DDAD = 6:77:6:5:6.
To make the model robust to different sensor setups, we apply a dynamic sampling strategy in each iteration:
\begin{enumerate}
    \item Randomly pick an image aspect ratio from [1.6, 3.3].
    \item Randomly choose the number of camera views from [2, 8].
    \item Calculate the maximum sequence length $T_{max}$ based on a hardware limit of 48 images per GPU.
    \item Randomly select a sequence length from [2, $T_{max}$] and set the batch size to fill the GPU memory.
\end{enumerate}
When finetuning \textbf{DVGT-2-NAVSIM} on NAVSIM, the input sequence from OpenScene is fixed to 8 views and 4 frames with an aspect ratio of 1.6, while the batch size is set to 1.

\section{Video Demonstration}
\cref{fig:demo} shows a sampled image from the video demo that demonstrates our model's predictions on the validation set. Given multi-view image sequences as input, \model performs robust geometry reconstruction and trajectory planning with high fidelity and consistency in an online manner, validating the effectiveness and efficiency of our VGA paradigm.

\end{document}